\documentclass[runningheads]{llncs}
\usepackage{tabularray}
\usepackage{graphicx}
\usepackage{amsmath,amssymb,amsfonts}
\usepackage{cite}
\usepackage{pdfpages}
\usepackage{epstopdf}
\usepackage{threeparttable}
\usepackage{amssymb}
\usepackage{multirow}
\usepackage{url}
\usepackage{booktabs}
\usepackage{animate}
\usepackage[misc]{ifsym}
\usepackage{tabularx}
\usepackage{bm}
\usepackage{indentfirst} 
\usepackage{gensymb}
\def\name{ColonNeRF}
\def\mynerf{DensiNet }

\begin{document}

\title{\name: High-Fidelity Neural Reconstruction of Long Colonoscopy}
%
%
\vspace{-7mm}
\author{Yufei Shi{$^{1*}$},~ Beijia Lu{$^{1*}$},~ Jia-Wei Liu{$^{12\dag}$},~ Ming Li{$^{2}$},~ Mike Zheng Shou$^{1\textrm{\Letter}}$\\\vspace{-8pt}{\small~}
\institute{{$^{1}$}Show Lab, {$^{2}$}Institute of Data Science,  National University of Singapore}}
\vspace{-7mm}
\renewcommand{\thefootnote}{\fnsymbol{footnote} }{}
\footnotetext[1]{Equal Contribution ~~ $^{\dag}$ Project Lead ~~  $^{\textrm{\Letter}}$ Corresponding Author}
%
%
%
\maketitle              

\begin{abstract}
Colonoscopy reconstruction is pivotal for diagnosing colorectal cancer. However, accurate long-sequence colonoscopy reconstruction faces three major challenges: (1) dissimilarity among segments of the colon due to its meandering and convoluted shape, (2) co-existence of simple and intricately folded geometry structures, and (3) sparse viewpoints due to constrained camera trajectories. To tackle these challenges, we introduce a new reconstruction framework based on the neural radiance field, ColonNeRF, for novel view synthesis of long-sequence colonos-copy. Specifically, ColonNeRF introduces a region division and integration module to reconstruct the entire colon piecewise, overcoming the challenges of shape dissimilarity. 
To learn both the simple and complex geometry in a unified framework,  ColonNeRF incorporates a multi-level fusion module that progressively models the colon structure.
Additionally, to eliminate the geometric ambiguities from sparse views, we devise a \mynerf  module for densifying camera poses under the guidance of semantic consistency.
We conduct extensive experiments on both synthetic and real-world datasets to evaluate our ColonNeRF. Quantitatively, ColonNeRF exhibits a 67\%-85\% increase in LPIPS-ALEX scores. Qualitatively, our reconstruction visualizations show much clearer textures and more accurate geometric details. These sufficiently demonstrate our superior performance over the state-of-the-art methods.

\end{abstract}

\section{Introduction}
Colorectal cancer (CRC) ranks as the fourth leading cause of cancer-related deaths \cite{b61}, yet early screening can elevate the 5-year survival rate to 90\% \cite{b62}. Therefore, identifying colorectal cancer in the early stage is essential\cite{b66,b67,b71}. Colonoscopy\cite{b64} has become one of the most crucial examinations for the early diagnosis of CRC due to its convenient operations and effectiveness. However, the diagnosis of CRC usually suffers from the complex colon structure and physicians potentially miss 22-28\% of polyps if solely relying on 2D scans\cite{b35}. These necessitate precise 3D colonoscopy reconstruction, which is also crucial for preoperative planning and medical training.

SLAM-based methods \cite{b60,b21,b23}  have been introduced into colonoscopy reconstruction by matching 2D image pixels and fusing geometries.
However, they perform poorly on novel view synthesis where a comprehensive understanding of 3D structures is required.  Consequently, they can not yield rich 2D scans of a colon, limiting their application in practice. Recently, to overcome this drawback, EndoNeRF \cite{b36} introduces NeRF\cite{b21} into medical scene reconstructions, focusing on surgical deformation tracking. 
A couple of other works also dedicate their energy to similar surgical scenes \cite{b36,b72,b73}.
In contrast, we aim at the precise reconstruction of a long-sequence colonoscopy whose intrinsic structures introduce several significant challenges. Firstly, the meandering and convoluted shape of a long colon results in heavy dissimilarity across different segments, posing obstacles for long-sequence reconstruction. Secondly, the co-existence of flat and intricately folded colon surfaces make it difficult for the model to correctly focus on concerning details. Lastly, the colonoscopy screening with constrained camera trajectories produces limited view images \cite{b36}, leading to geometric ambiguities for reconstruction \cite{b21}.


To resolve the above-mentioned challenges, we propose a new model for 3D colonoscopy reconstruction dubbed ColonNeRF that comprises three main modules. To overcome the challenges of dissimilarity among colon regions, we design a region division and integration module to represent the long colon. Specifically, the division module first divides the colon into multiple segments based on its curvature in nature. In each segment, we design a multi-level fusion module to progressively model the colon textures and geometric details in a coarse-to-fine way. Then, our integration module fuses the divided segments with two filtering strategies. Additionally, to recover the geometric details missing in sparse viewpoints, we present a DensiNet module that encourages our model to learn colon features from extra angles, i.e., original pose, spinning around pose, and helix rotating pose, under the regularization of DINO-ViT semantic consistency in each stage. 
We conduct extensive experiments on synthetic and real-world datasets to verify our ColonNeRF, demonstrating its superior performances over existing methods.


\section{Method}
\subsection{Preliminaries of Neural Radiance Fields}
Neural Radiance Fields (NeRF) \cite{b1} synthesize novel views of a scene by mapping 5D coordinates, comprising 3D position $\textbf{x}$ and 2D viewing direction $\textbf{d}$ to RGB color $\textbf{c}$ and volumetric density $\sigma$. Each pixel in an image corresponds to a ray $\textbf{r}(\tau)=\textbf{o}+\tau \textbf{d}$, where $\textbf{o}$ is the camera origin, and $\textbf{d}$ is the ray direction, $\tau$ is the distance between the origin point and sample point. The predicted color $\textbf{C}(\textbf{r})$ of the pixel can be represented as:
{
\begin{equation}
\begin{array}{c}
\textbf{C}(\textbf{r})=\int_{\tau_{\text {near }}}^{\tau_{\text {far }}} T(\tau) \sigma(\textbf{r}(\tau)) \textbf{c}(\textbf{r}(\tau), \textbf{d}) d \tau\end{array}, \,
T(\tau)=\exp \left(-\int_{\tau_{\text {near }}}^{\tau} \sigma(\textbf{r}(s)) d s\right).
\end{equation}}NeRF \cite{b1} model optimizes the radiance field by minimizing the mean squared error between the synthetically rendered color and the ground truth color:
{
\begin{equation}
\begin{array}{c}
\mathcal{L}_{\mathrm{pixel}}=\sum_{\textbf{r} \in R_{i}}\|(\textbf{C}(\textbf{r})-\hat{\textbf{C}}(\textbf{r}))\|^{2},
\end{array}
\end{equation}}where $R_i$ is the set of input rays during training, $\hat{\textbf{C}}(\textbf{r})$ and $\textbf{C}(\textbf{r})$ is the ground truth and predicted RGB colors for ray \textbf{r}.

\subsection{Region Division Module}
\label{subsec:B}

To address the inherent dissimilarities in different colon segments that are characterized by varying diameters and curvatures, we develop a region division module for the colon's meandering and convoluted structure. 
Specifically, it segments the colon into blocks at bends or locations with significant angle changes. This approach promotes the overall quality and accuracy of the reconstruction.

We adapt this segmentation strategy to suit each dataset's specific geometry characteristics and ensure a 30\% overlap between adjacent blocks to maintain seamless transitions. This overlapping strategy is illustrated in Fig. 1,  where each block is represented with a central red region surrounded by two orange regions, indicating the areas of overlap. This methodological approach, detailed further in our ablation study, ensures a more accurate reconstruction of the colon complex geometry.

\subsection{Multi-Level Fusion Module}

The multi-level fusion module initiates with inputs of low sparsity RGB, depth, and pose data. It progressively incorporates denser data, enabling a smooth transition from coarse to fine details, thus enhancing the effectiveness of the feature extraction process.  The level of data sparsity at each $i$th stage of the input model is calculated using the formula $\frac{2^{n}}{F*2^{i}}$, where $i$  denotes the stage number, and $F$ represents the total frames during the detection. Each stage of the module includes two sub-modules: \mynerf and the visibility module.  \mynerf generates RGB and density $\sigma$ values for each spatial position. The visibility module calculates the transparency $T_{i}$ for each spatial ray and supervises transparency with the density $\sigma$ output from DensiNet, following the formula to calculate the transparency loss:
$\mathcal{L}_{\text{trans}} = \parallel T_\mathrm{i} - \sigma_\mathrm{i} \parallel$.
As the model progresses, it inherits the parameters of \mynerf and the visibility module from the previous stage, adding two residual connections to link the color and density outputs from the previous stage to the next. The final output combines the newly calculated RGB $c_2$ and density $\sigma_2$ values with outputs from each stage, resulting in a comprehensive final image.
\begin{equation}
\sigma_{\mathrm{output}}  =\sigma_\mathrm{L}({\textstyle \sum_{\sigma_{1} }^{\sigma_{\mathrm{n}}}} ), \quad \quad C_{\mathrm{output}}  = \zeta({\textstyle \sum_{\mathrm{C}_{1} }^{C_{\mathrm{n}}}}).
\end{equation}
The activation functions applied to the final values of $\sigma$ and $c$ include the Sigmoid function $\sigma_\mathrm{L}$ for density and a Softplus function $\zeta$ for color. 
\begin{figure}[t]
\centering
\includegraphics[width=\textwidth]{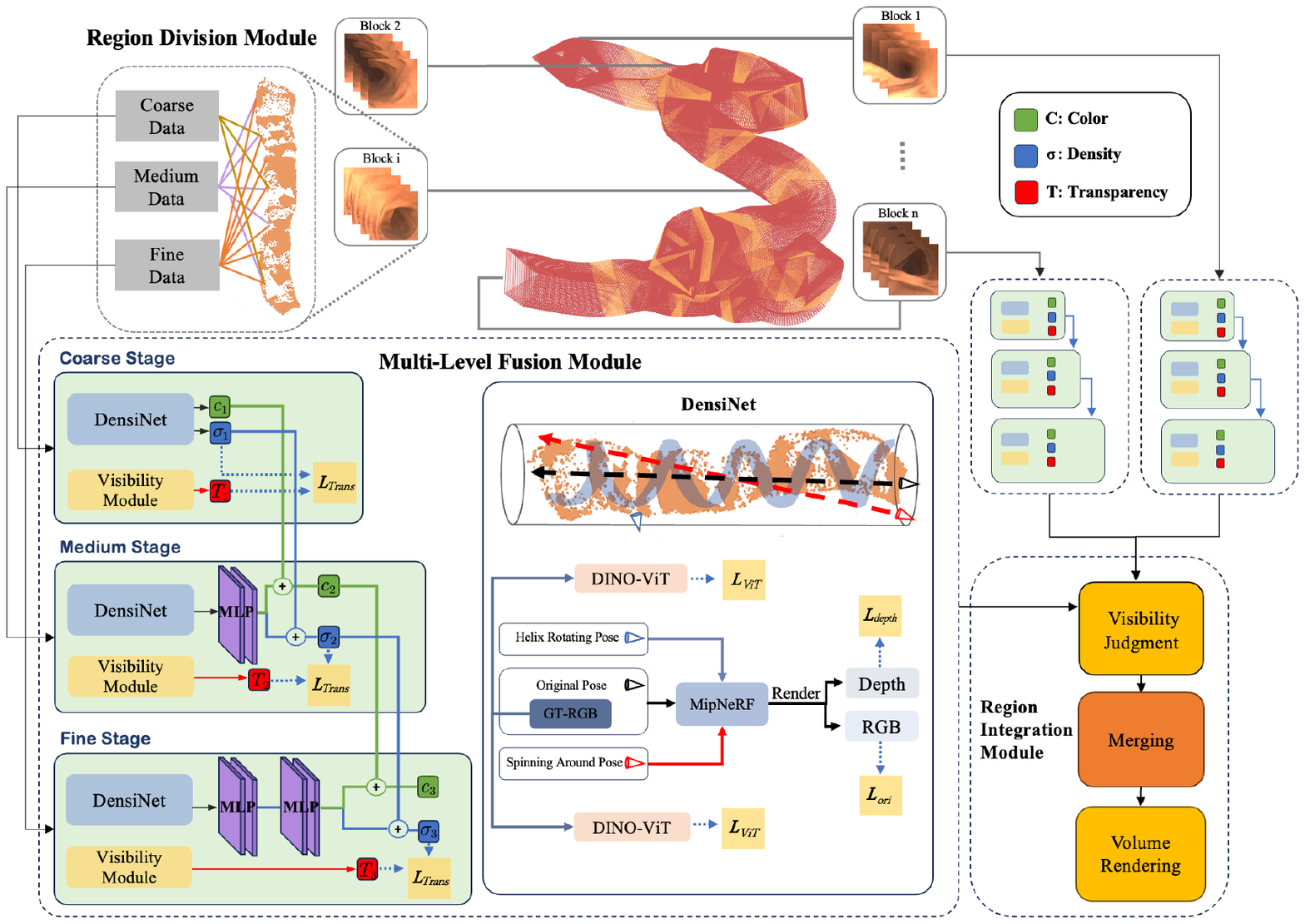}
\caption{Illustration of our proposed novel approach of neural rendering for colonoscopy. 
}
\label{Overview}
\end{figure}
\subsection{\mynerf Module}
To deal with the challenge of sparse camera viewpoints, we design the \mynerf module, which leverages MipNeRF \cite{b3} as its backbone. \mynerf module enhances model's ability to capture the colon features through three angles.

\noindent\textbf{Original Pose.} To learn the structure from the original viewpoint, we design the original pose loss. The original pose loss is the sum of the difference between extracted patches and their counterparts in the post-rendering images and the MSE loss between
sampled points and their corresponding post-rendered points:
$\mathcal{L}\mathrm{_{patch}}  = \mathcal{L}_\mathrm{p}(\bm{C}_\mathrm{p}, f(\bm{C}_\mathrm{p}))+\mathcal{L}_\mathrm{p}(\bm{D}_\mathrm{p}, f(\bm{D}_\mathrm{p}))$ and 
$\mathcal{L}\mathrm{_{rand}}  = \mathcal{L}_\mathrm{m}(\bm{C}_\mathrm{s}, f(\bm{C}_\mathrm{s}))+\mathcal{L}_\mathrm{m}(\bm{D}_\mathrm{s}, f(\bm{D}_\mathrm{s}))$, 
 where $\mathcal{L}_\mathrm{p}$ represents the patch loss. $\bm{C}_\mathrm{p}$ and $\bm{D}_\mathrm{p}$ represent the sampled points of RGB and depth image obtained in the original view. $f(\bm{C}_\mathrm{1})$ and $f(\bm{D}_\mathrm{1})$  refer to the RGB and depth output results from the MipNeRF \cite{b3}. $\mathcal{L}_\mathrm{m}$ represents the MSE loss. The variables $\bm{C}_\mathrm{s}$ and $\bm{D}_\mathrm{s}$ correspond to the points sampled from the RGB and depth images using a random selection strategy. And we could get the final original pose loss:
$\mathcal{L}\mathrm{_{ori}}=\mathcal{L}\mathrm{_{patch}}+\mathcal{L}\mathrm{_{rand}}$.

\noindent \textbf{Spinning Around Pose.} To enhance the reconstruction of geometric structures around the original pose, we employ a rotation transformation to obtain spinning around pose. For any given pixel $\bm{P}(x_{\mathrm{i}}, y_{\mathrm{i}})$ on the original view, its corresponding position on the destination pose $\bm{P}_{\mathrm{des}}$ can be represented as:
{
\begin{equation}
\begin{array}{c}
\bm{P}_\mathrm{des}   = \begin{bmatrix}
  \bm{R}_\mathrm{des}& \bm{t}_\mathrm{des}  \\
  \bm{0} & \bm{1}
\end{bmatrix} \begin{bmatrix}
 \bm{R}_\mathrm{ori}& \bm{t}_\mathrm{ori} \\
 \bm{0} & \bm{1}
\end{bmatrix}^{-1}\cdot \bm{D}\cdot {\bm{P}_\mathrm{ori}}\end{array}.
\end{equation}}

In this formula, $\bm{R}_\mathrm{des}$ and $\bm{t}_\mathrm{des}$ denote the rotation matrix and translation vector for the destination pose. Similarly, $\bm{R}_\mathrm{ori}$ and $\bm{t}_\mathrm{ori}$ represent those of the original pose. $\bm{D}$ is used to convert pixel coordinates $\bm{P}(x_{i}, y_{i})$ to camera world coordinates ($x$, $y$, $z$). 
We carry out rotational sampling around the initial original pose, rotating along the $x$, $y$, and $z$ axes at different angles ($5^\circ$ , $2.5^\circ$, and $1.25^\circ$) to generate $216$ directional poses. We integrate all the rays from the 216 poses, randomly selecting 3,136 rays each time as our spinning around pose.

 \noindent\textbf{Helix Rotating Pose.} Due to the spiral characteristics of colon folds, the \mynerf module adopts a spiral-shaped sampling trajectory to capture the 3D structure of the folds. We interpolate between the current pose $\bm{P}_\mathrm{3}$ ($x_{3}, y_{3}, z_{3}$) and neighboring pose $\bm{P}_\mathrm{4}$ ($x_{4}, y_{4}, z_{4}$) using the Slerp (Spherical Linear Interpolation) algorithm, which yields a quaternion representing the direction at the intermediate position. Through image warping, we obtain the depth and RGB images in many unseen views, which serve as the pseudo ground truth label. To supervise the colon geometry structure in the rotated view, we compute the discrepancy between these target depths and the depths rendered by the \mynerf under the same poses using the following loss function:
{
\begin{align}
\mathcal{L}\mathrm{_{depth}}  = \mathcal{L}_\mathrm{1}(\bm{H}_\mathrm{d}, \bm{D}_\mathrm{3})+\mathcal{L}_\mathrm{1}(\bm{S}_\mathrm{d}, \bm{D}_\mathrm{3}),
\end{align}}where $\mathcal{L}\mathrm{_1}$ represents the Smooth L1 Loss \cite{b55}. $\bm{H}_\mathrm{d}$  and $\bm{S}_\mathrm{d}$ denote the depth obtained from the helix and spin transformation, and $\bm{D}_\mathrm{3}$ corresponds to the depth rendering result for the corresponding transformation method. We use the MSE loss to calculate the loss between the extracted features:
{
\begin{equation}
\mathcal{L}\mathrm{_{ViT}}  =\mathcal{L}_\mathrm{m}( F_\mathrm{ViT}(C_\mathrm{o}), F_\mathrm{ViT}(f(C_\mathrm{r}))).
\end{equation}}Here, $F_\mathrm{ViT}$ represents the pre-trained model that we employ to extract semantic information from the RGB of the original views $C_\mathrm{o}$ and the rendering RGB results in the rotated views $f(C_\mathrm{r})$.
 

 \begin{table}
\centering
\caption{ Quantitative evaluation and ablation study about different state.}
\resizebox{0.8 \linewidth}{!}{
\begin{tblr}{
  colsep = 0.8pt,
  rowsep =0.3pt,
  cells = {c},
  cell{2}{1} = {r=2}{},
  cell{4}{1} = {r=2}{},
  cell{6}{1} = {r=2}{},
  cell{8}{1} = {r=2}{},
  cell{10}{1} = {r=2}{},
  cell{13}{1} = {r=4}{},
  cell{13}{2} = {r=2}{},
  cell{15}{2} = {r=2}{},
  cell{17}{1} = {r=4}{},
  cell{17}{2} = {r=2}{},
  cell{19}{2} = {r=2}{},
  cell{21}{1} = {r=4}{},
  cell{21}{2} = {r=2}{},
  cell{23}{2} = {r=2}{},
  cell{25}{1} = {r=2}{},
  cell{25}{2} = {r=2}{},
  hline{1-2,12-13,17,21,25,27} = {-}{},
}
Qualitative~                             & Datasets                         & D-MSE  & PSNR↑          & VGG↓                     & ALEX↓                    & MS-SSIM↑                 \\
NeRF                                     & Syn                              &   3.37       & 26.10          & 0.4888                   & 0.4405                   & 0.8266                   \\
                                         & Real                             &   14.61       & 25.86          & 0.4273                   & 0.3745                   & 0.8536                   \\
MipNeRF                                  & Syn                              &   18.06       & 24.96          & 0.4863                   & 0.4367                   & 0.7954                   \\
                                         & Real                             &   10.50       & 23.29          & 0.4142                   & 0.3470                   & 0.7702                   \\
FreeNeRF                                 & Syn                              &   2.71       & 24.80          & 0.5141                   & 0.4815                   & 0.7881                   \\
                                         & Real                             &   13.33       & 25.16          & 0.4096                   & 0.3473                   & 0.8396                   \\
EndoNeRF                                 & Syn                              &   1.08       & 21.67          & 0.4985                   & 0.4378                   & 0.6934                   \\
                                         & Real                             &    10.25      & 21.62          & 0.5077                   & 0.4889                   & 0.7061                   \\
ColonNeRF                                & Syn                              &   0.07       & \textbf{26.70} & \textbf{\textbf{0.3989}} & \textbf{\textbf{0.2605}} & \textbf{\textbf{0.8373}} \\
                                         & Real                             &   0.21       & \textbf{25.54} & \textbf{\textbf{0.4001}} & \textbf{\textbf{0.3259}} & \textbf{\textbf{0.8676}} \\
Ablation Study                           & Ablation State                   & Datasets & ~PSNR↑         & VGG↓                     & ALEX↓                    & MS-SSIM↑                 \\
{Multi-level \\fusion Module   }         & {Coarse}                      & Syn      & 25.28          & 0.4228                   & 0.2993                   & 0.7986                   \\
                                         &                                  & Real     & 24.97          & 0.4242                   & 0.3393                   & 0.8244                   \\
                                         & Medium                           & Syn      & 25.94          & 0.4097                   & 0.2770                   & 0.8176                   \\
                                         &                                  & Real     & 25.47          & 0.4143                   & 0.3298                   & 0.8474                   \\
{~Division and\\Integration Module}      & w/o Division                     & Syn      & 20.18          & 0.5883                   & 0.5639                   & 0.6743                   \\
                                         &                                  & Real     & 24.49          & 0.4467                   & 0.3731                   & 0.7944                   \\
                                         & w/o Integration                  & Syn      & 26.62          & 0.4014                   & 0.2620                   & 0.8344                   \\
                                         &                                  & Real     & 25.88          & 0.4016                   & 0.3288                   & 0.8655                   \\
{Different View\\as input \\in DensiNet} & {1 View}                      & Syn      & 25.05          & 0.4407                   & 0.3798                   & 0.8004                   \\
                                         &                                  & Real     & 25.47          & 0.4254                   & 0.4031                   & 0.8523                   \\
                                         & {2 View}                      & Syn      & 25.79          & 0.4086                   & 0.2692                   & 0.8101                   \\
                                         &                                  & Real     & 25.77          & 0.4128                   & 0.3782                   & 0.8533                   \\
                                         & {Full Model\\(Fine) (3 View)} & Syn      & \textbf{26.70} & \textbf{0.3989}~         & \textbf{0.2605}          & \textbf{0.8373}          \\
                                         &                                  & Real     & \textbf{25.96} & \textbf{0.4001}          & \textbf{0.3259}          & \textbf{0.8676}          
\end{tblr}
}
\label{ablation}
\end{table}
\subsection{Region Integration Module}

 \noindent\textbf{Filtering Method.} To enhance the efficiency of the colon fusion process, we establish two mechanisms for filtering useless blocks. Firstly, we calculate the Euclidean distance between the observation points and the line connecting the centers of two adjacent blocks. A block is retained if this distance is less than the threshold. The second filtering strategy leverages the visibility module to calculate the transparency of this point to the respective block. If the transparency falls below the threshold, we exclude that block.
 
\noindent\textbf{Merging Method.}
To merge adjacent segments after a dual filtering strategy, we employ the Inverse Distance Weighting (IDW) technique \cite{b32} because of its effectiveness in realizing a smooth transition between adjacent segments. The merging weight $W_i$ of block i for interpolation is according to the formula:
$ W=\parallel \text{center}, P_\mathrm{t}\parallel^{-\varepsilon }$,
where $\varepsilon$ denotes the rendering blend ratio. And we normalize $W_i$ to obtain the weight $w_i$. We will use a weighted sum of the results under different blocks to get the final RGB and Depth. \noindent{Final loss function is shown as follows:}
\begin{equation}
\mathcal{L}\mathrm{_{all}} =\mathrm{\lambda_1}\mathcal{L}_{depth} + \lambda_2\mathcal{L}_{ori} +\\
\mathrm{ \lambda_3}\mathcal{L}_{ViT}+\lambda_4\mathcal{L}_{trans},
\end{equation}
where $\lambda_1$, $\lambda_2$, $\lambda_3$, and $\lambda_4$ represent the weights of different losses.

\section{Experiments}
\subsection{Datasets and Implementation Details}
We utilize synthetic (SimCol-to-3D 2022 \cite{b22}) and real-world (C3VD Descending Colon datasets \cite{b34}) colon phantom datasets for evaluation. The synthetic dataset comprises 989 frames. We approximately sample one out of four frames as test frames and use the remaining frames for training, resulting in $233$ test images and $756$ train images. The real-world dataset is similarly divided into $35$ train images and $19$ test images. We set $\lambda_1$, $\lambda_2$, $\lambda_3$, and $\lambda_4$ to $8$, $1$, $10$, and $1$, respectively. We adopt five metrics to evaluate novel views and depths: PSNR, LPIPS-VGG, LPIPS-ALEX, MS-SSIM, and Depth-MSE.

\subsection{Comparison with State-of-the-Art Methods}
We compare our model with NeRF \cite{b1}, MipNeRF \cite{b3}, FreeNeRF \cite{b38}, and EndoNeRF \cite{b36} on the synthetic \cite{b22} and real-world \cite{b34} datasets . 


\begin{figure}[t]
\centering
\includegraphics[width=3.5in,angle=90]{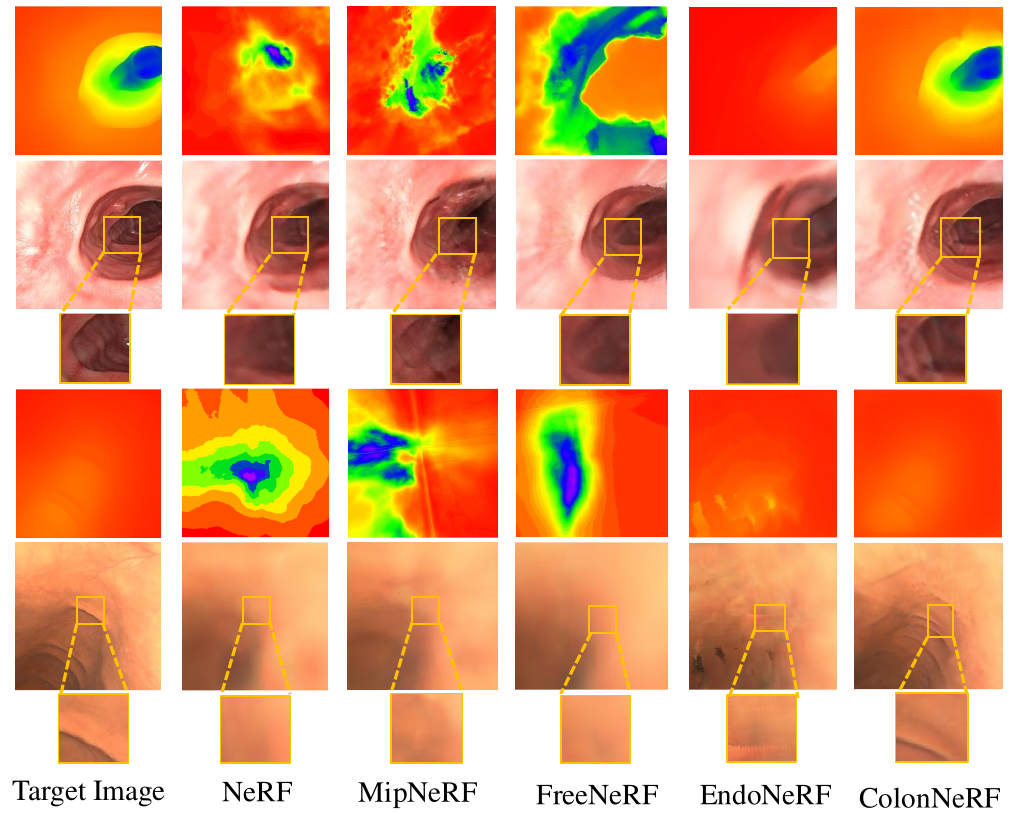}
\caption{Novel view synthesis RGB and corresponding depth results of different methods on the synthetic dataset. }
\label{sota}
\end{figure}

\noindent\textbf{Qualitative Comparison.} As shown in Fig. \ref{sota}, our novel view synthesis results on both synthetic and real-world datasets demonstrate significant improvements over baselines. Baseline methods produce noticeable blurs with missing critical details such as the folds structure. Moreover, the depth maps from baselines show significant deviations from the ground truth. In contrast, our method renders high-quality novel views and depths with finer details.


 \noindent\textbf{Quantitative Comparison.} We report quantitative comparisons in Tab. \ref{ablation}. The labels `Syn' and `Real' correspond to results of the SimCol-to-3D \cite{b22} and C3VD datasets \cite{b34}. Our model demonstrates the highest quantitative performance over all metrics. Specially, our method achieves large improvement of 67\% against other methods in terms of LPIPS-ALEX, and renders much better depth maps ($15.4 \times$ and $48.8 \times$ better for depth-MSE on `Syn' and `Real' datasets). These demonstrate the superiority of ColonNeRF.

\begin{figure}[t]
\centering
\includegraphics[width=\textwidth]{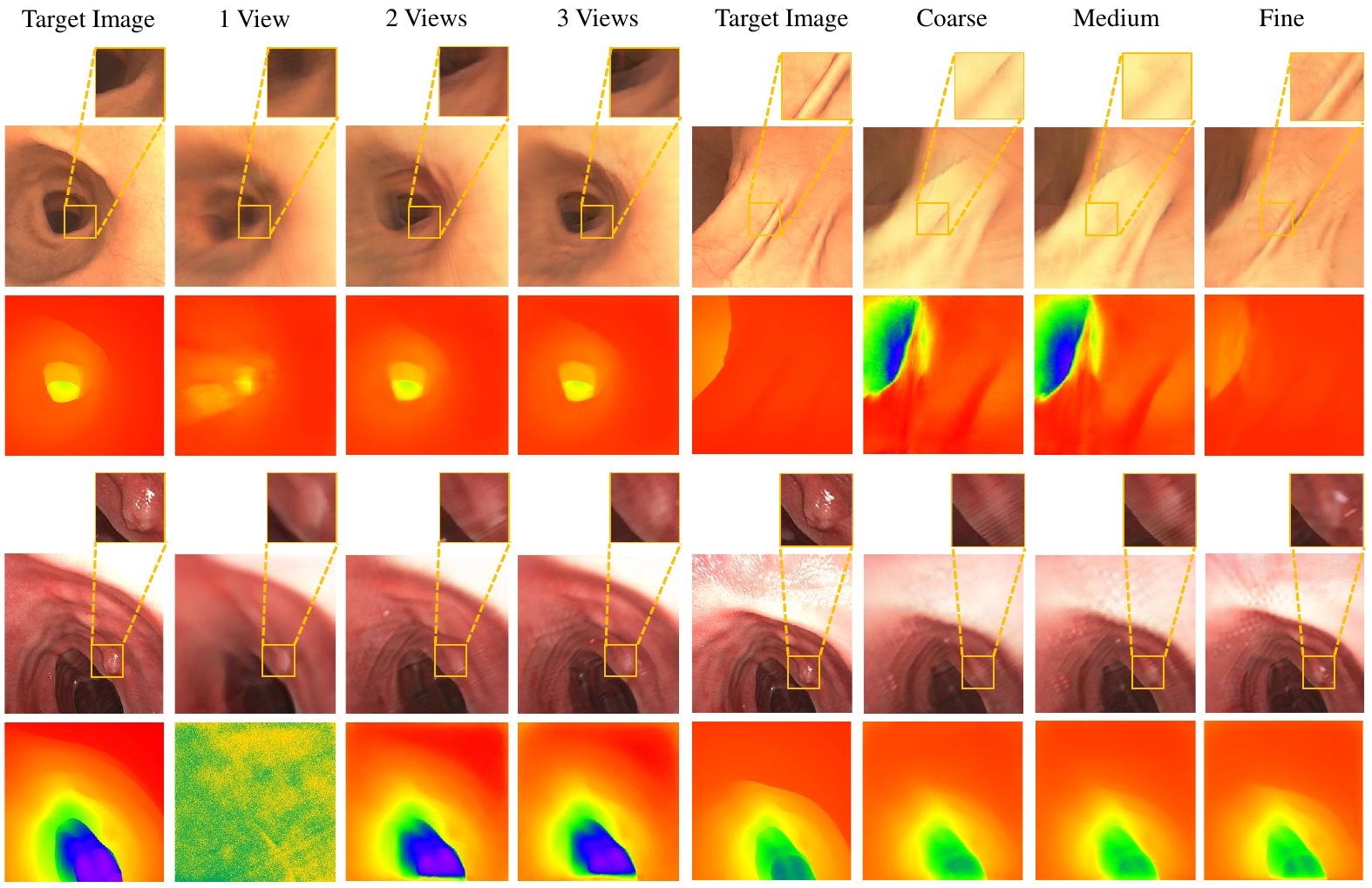}
\caption{Ablation Study about different views as input in DensiNet module and different stages in the multi-level fusion module.}
\label{Multi_level}
\end{figure}

\subsection{Ablation Study}
 \noindent\textbf{Effects of Multi-Level Fusion Module.}
 We evaluate each processing stage – coarse, medium, and fine. When the model operates with only the coarse stage, we input the $c_\mathrm{1}$ and $\sigma_\mathrm{1}$ from the coarse stage directly into the subsequent integration module. As the figures indicate, only coarse results in noticeably blurred reconstructions, particularly around the edges. With the incremental addition of stages, the model shows a more comprehensive depiction of detailed information with less noise as shown in Fig. \ref{Multi_level} and Tab. \ref{ablation}. Considering efficiency and computational time, we ultimately choose to implement three stages. 
 
 \noindent\textbf{Effects of Division and Integration Module.}
We present the results in supplementary and Tab. \ref{ablation}. Without the division module, a single block for processing all intestinal data results in noticeable distortions and artifacts. Because it is difficult to handle the varied appearance and drastic angle changes in the meandering and convoluted colon. 
The integration module significantly improves the reconstruction outcomes at transitions between adjacent block regions. 

 \noindent\textbf{Effects of \mynerf Module.}
We explore the impact of integration inputs from different poses. As depicted in Fig. \ref{Multi_level} and Tab. \ref{ablation}, 1 view: original pose, 2 views: original pose + helix rotating pose, 3 views: original pose + helix rotating pose + spinning around pose. 
Our empirical evidence shows that incorporating each new viewpoint provides guidance about semantic consistency and improves the accuracy in depth estimation and the overall clarity of the rendering images.

\section{Conclusion}
We introduced ColonNeRF, an innovative framework designed for long-sequence colonoscopy reconstruction. To tackle the challenges of such a task, we proposed a region division and integration module to segment long-sequence colons into short blocks, a multi-level fusion module to progressively model the block colons from coarse to fine, and a \mynerf module to densify the sampled camera poses under the guidance of semantic consistency. Extensive experiments on synthetic and real-world data demonstrate the superiority of ColonNeRF.

\bibliographystyle{IEEEtran}
\bibliography{ref}

\includepdf[pages=-]{supp.pdf}
\end{document}